\documentclass[10pt,twocolumn,letterpaper]{article}

\usepackage{cvpr}
\usepackage{times}
\usepackage{epsfig}
\usepackage{graphicx}
\usepackage{amsmath}
\usepackage{amssymb}

\usepackage[pagebackref=true,breaklinks=true,letterpaper=true,colorlinks,bookmarks=false]{hyperref}

\cvprfinalcopy 

\def\cvprPaperID{2975} 

\ifcvprfinal\fi
\begin{document}

\title{Personalization of Saliency Estimation}

\author{Bingqing Yu\\
Centre for Intelligent Machines, McGill University\\
{\tt\small bingqing.yu@mail.mcgill.ca}
\and
James J. Clark\\
Centre for Intelligent Machines, McGill University\\
{\tt\small clark@cim.mcgill.ca}
}
\maketitle

\begin{abstract}
Most existing saliency models use low-level features or task descriptions when generating attention predictions. However, the link between observer characteristics and gaze patterns is rarely investigated. We present a novel saliency prediction technique which takes viewers' identities and personal traits into consideration when modeling human attention. Instead of only computing image salience for average observers, we consider the interpersonal variation in the viewing behaviors of observers with different personal traits and backgrounds. We present an enriched derivative of the GAN network, which is able to generate personalized saliency predictions when fed with image stimuli and specific information about the observer. Our model contains a generator which generates grayscale saliency heat maps based on the image and an observer label. The generator is paired with an adversarial discriminator which learns to distinguish generated salience from ground truth salience. The discriminator also has the observer label as an input, which contributes to the personalization ability of our approach. We evaluate the performance of our personalized salience model by comparison with a benchmark model along with other un-personalized predictions, and illustrate improvements in prediction accuracy for all tested observer groups.
\end{abstract}

\section{Introduction}

Humans have developed the ability to select target areas to look at given a visual scene. When confronted by a clustered visual stimulus, people tend to direct their eye movements towards the locations that attract their attention the most. In this way, the most relevant objects of an image are brought to the central part of the retina, and it becomes feasible for the human brain to process and interpret detailed scenes in real time. Therefore, the development of computational models of visual attention have attracted much interest. Numerous techniques have been developed with the goal of understanding, modeling and predicting where people look at in visual scenes, as this capability may find applications in a wide range of real-world tasks, such as advertising design, product marketing, surveillance, real-time robotic vision systems and human-machine interaction. Most basic attention models use low-level features such as orientation and illuminance in constructing image salience predictions \cite{itti1998model}. Koch and Ullman's feed-forward neural model was the pioneering work in attention modeling \cite{koch1987shifts}, and many other models have been developed since then. In particular, one important viewpoint, which is the top-down modulation of attention, has been explored by many researchers \cite{einha2008task}. Itti and Koch proposed an attention model that modulates low-level feature maps according to the visual-task \cite{itti2001feature}. This two-component framework, which includes the bottom-up and top-down attention, has inspired many developments in the field of saliency map computation. However, recent publications in the field of neuroscience have suggested that reward plays an important role in the selection of attention. Even though early work in the field of attention modeling were based on a framework including only top-down and bottom-up factors, Chelazzi \etal proposed in their work that the reward-associated or value-associated stimuli that people have experienced previously may have a lasting and generalized impact on the human attention system \cite{chelazzi2014altering}. Christian \etal took this idea a step further by proposing a three-component attentional model, which combines low-level features, task-specific motivations and value driven priorities \cite{klink2014priority}. By rewarding their participants for selecting certain targets during the training phase of their experiment and observing the subsequent behaviors of the participants in the testing phase, Chelazzi and colleagues showed that human attention can be shaped significantly via reward-based learning. In other words, where people are more likely to look at can be affected by the value judgements that they learned from past experience. This novel concept of value modulation of attention is missing in traditional saliency models, which include only low-level feature-based and task-based components. The lack of consideration for value-driven factors is a shortcoming for current saliency models, which focus only on the visual stimuli and tasks when analyzing attention. However, in general, people have specific individual characteristics and personal values which influence their attentional allocation in ways that differ from person to person. Clearly, the traditional saliency models that ignore the role of value-driven attentional processing in the human brain are insufficient to accurately predict salience for a given individual, as these models are not able to take individual variations into consideration and generate personalized saliency predictions. Thus, our goal is to construct an attention model that can compute image saliency taking individual variations into consideration. Recent work \cite{xu2017beyond} has attempted to use deep learning methods to make personalized saliency predictions, however, our proposed method has some significant advantages. First, no extra data generation step is required by our method. All of the eye fixation data that we use for training are taken from pre-existing datasets. We do not need to purposely construct a modified dataset for saliency personalization tasks. It is possible in our approach to apply  the trained network by replacing the training eye fixation data with any other eye fixation predictions generated from any state-of-the-art saliency model to generate personalized saliency maps. In addition, we only use a simple encoding method to represent viewer identity information during training without needint to conduct a detailed survey for information collection. Last but not least, our method is effective for performing prediction tasks in different contexts. Not only it can generate personalized gaze predictions for natural image stimuli, it can also generate personalized gaze predictions given images containing text-only content. 

\begin{figure}[t]
\begin{center}
\fbox{
   \includegraphics[width=0.9\linewidth]{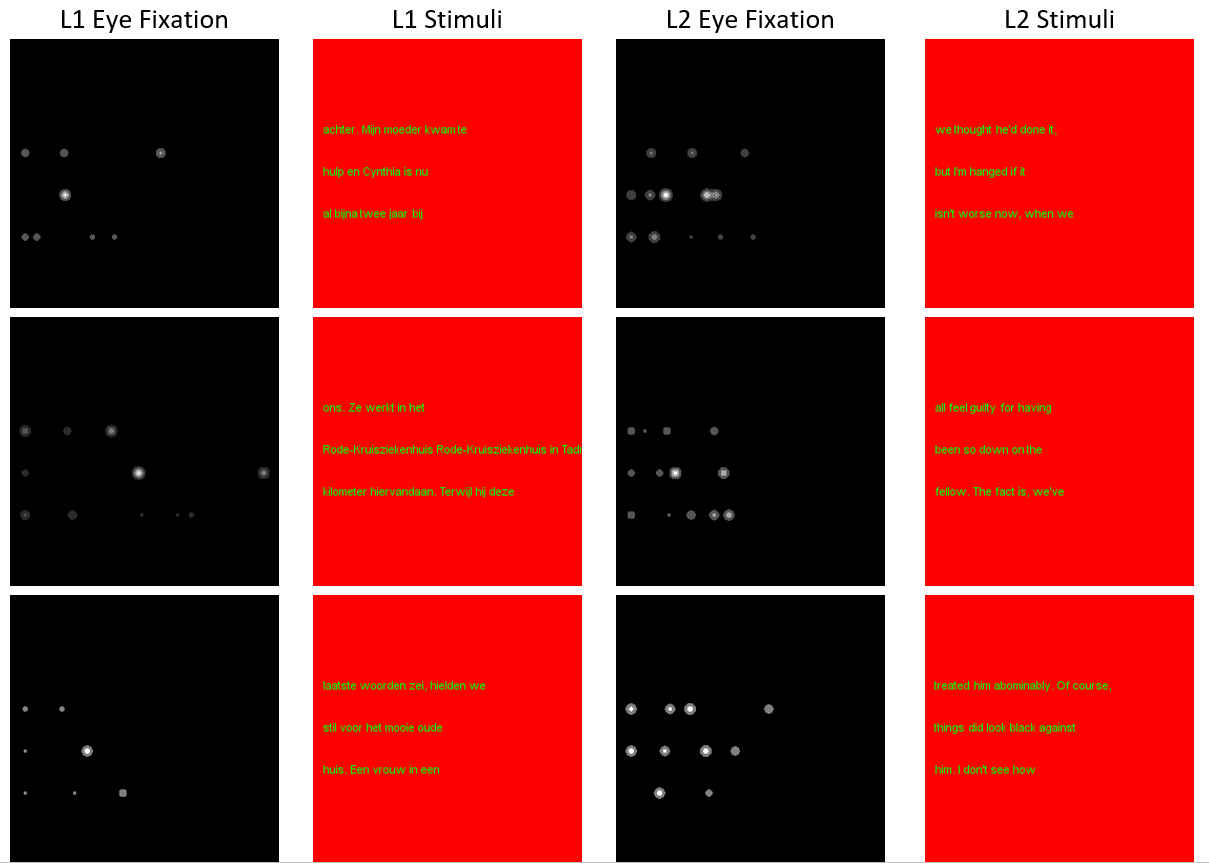}}
\end{center}
   \caption{Comparison of eye tracking data collected during mother tongue (L1) reading versus second language (L2) reading. Starting from the left, the first column shows the eye fixation heat maps recorded for L1, with the corresponding reading material presented in the second column. The third column shows the eye fixation heat maps generated for L2, and the corresponding reading material is presented in the fourth column.}
\label{fig:eyetotextgt}
\end{figure}

We demonstrate our proposed method on two quite different datasets. One dataset provides eye tracking data of viewers belonging to different age groups \cite{accik2010developmental}, looking at images of natural scenes. By comparing the performance of older adults to the performance of young children, scientists using this dataset were able to confirm that older adults display a more explorative viewing performance than do children. Their fixations are more spread-out over each image stimulus, and their saccades are of lower amplitude compared to the performance of younger children. The other dataset, named GECO, the Ghent Eye-Tracking Corpus, also demonstrates a difference in viewing behaviors of two groups of participants \cite{cop2017presenting}. The eye-tracking data was collected by hiring Dutch-English bilinguals and asking them to read half of a novel in their first language, and the other half in their second language. The results showed that the difference in eye fixation patterns from mother tongue to non-mother tongue reading is very similar to the changes in reading patterns from child to adult reading. More specifically, people display more fixations and shorter saccades when changing to non-mother tongue reading. Figure~\ref{fig:eyetotextgt} shows some ground truth eye fixation heat maps generated from the GECO dataset, along with the corresponding reading material. It can be observed that, in general, when people are reading in their mother tongue (L1), the ground truth eye fixation heat maps display much less fixation points and fixation durations compared to the L2 case where people are reading in their second language.

These findings strongly suggest that attention can be modulated by observers' backgrounds and past experiences.

\subsection{Proposed Method}

\begin{figure}[t]
\begin{center}
\fbox{
   \includegraphics[width=0.9\linewidth]{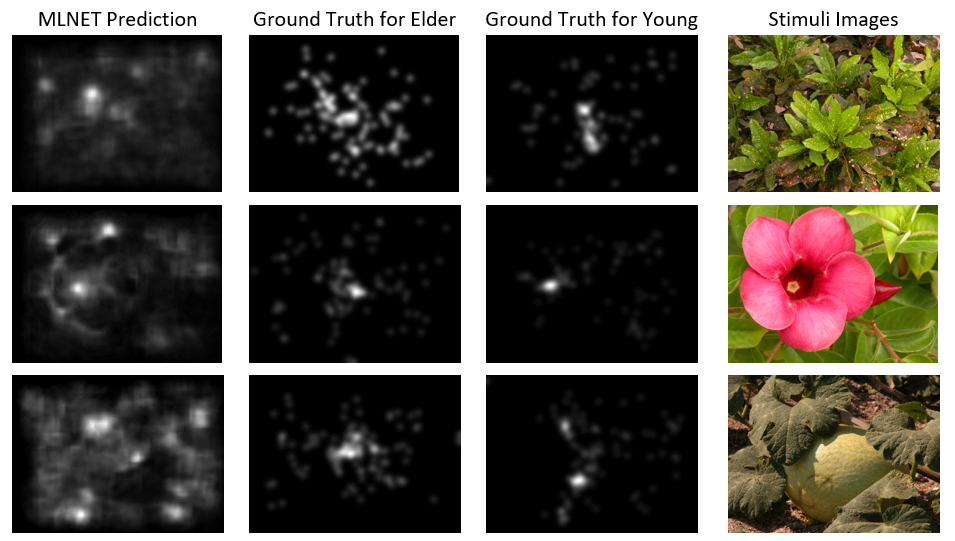}}
\end{center}
   \caption{From the left, the first column shows examples of MLNET saliency predictions. The second column is the ground truth eye fixation heat maps for the elder viewer group. The third column is the ground truth eye fixation heat maps for the younger viewer group. The corresponding image stimuli are listed in the fourth column.}
\label{fig:person_mlnetintro}
\end{figure}

Our proposed method is designed to model the variations of viewing behaviors displayed among different groups of observers. When viewers are looking at similar visual stimuli, we hypothesize that their attention allocation is modulated by their subconscious value-driven signals developed through their past experiences and cultural biases. As a consequence, the eye fixation heat maps depend on not only the image stimuli, but also distinguishing information of each individual viewer. For example, for a certain group of people, some locations in a scene may get high saliency whereas they may be unattended in another group of people. Thus, we assume that, in attempting to produce highly accurate saliency predictions, a saliency model needs to consider both the displayed visual stimuli and various characteristics of viewers. Figure~\ref{fig:person_mlnetintro} shows some examples of saliency predictions generated by MLNET, a state-of-the-art saliency model (according to the MIT saliency benchmark). This network is trained over a large population of viewers, of different ages. However, when compared with the ground truth fixation heat maps for younger children from the Age Study dataset, it can be seen clearly that MLNET performs poorly in generating gaze patterns that represent the viewing characteristics of younger children observers. To improve on the prediction capability of existing saliency methods, we need to incorporate viewers' identity information into our model and use it to fine-tune and provide personalized saliency predictions.

Although creating a fresh saliency model from scratch, training only on the population subset of interest, is an option, we choose to use and benefit from a pre-trained neural networks when designing our model. This has the advantage of saving on training time, and the ability to use a larger training set from a large population. However, we need to have some way of augmenting this pre-trained network to provide the personalization. To do this modification, we feed the output of the pre-trained network into a modified conditional Generative Adversarial Network (GAN) (see Section~\ref{relatedwork}). This produces a network that is able to generate personalized saliency predictions given visual stimuli and viewers' identities. Our augmented network contains a generator which performs an image translation task. It takes the input image pixels at one end, and produces, at the other end, an output image corresponding to the desired saliency prediction. Using the remarkable concept of the traditional GAN, our generator is combined with a discriminator. The discriminator's task is to distinguish fake salience maps from real salience maps. The generator will later sense the answer backpropagated from the discriminator and become better at generating appropriate saliency maps. The network has multiple inputs, the first of which is the RGB image that we give to the input layer of the generator. For the example of the Age Study dataset, the input RGB image uses one channel to encode a grayscale version of the displayed image stimulus and uses another channel to encode a heat map containing (during training) ground truth heat maps or (during testing) heat maps generated by the ML-NET salience algorithm, for image data collected from both the younger and elder viewer groups. in the case of the GECO dataset, the input RGB image uses one channel to encode a grayscale image containing an image of a page of printed text corresponding to the part of novel read by the participants, and uses another channel to represent gaze fixations assuming people look at each word uniformly and independently of the readers' identities and reading habits. The label information that indicates the group that each viewer belongs to (e.g. young or old, or native language vs. second language) is encoded as a tensor. This information is injected into an interior layer of the generator and discriminator. The discriminator, built following the generator, not only takes as input candidate salience map to discriminate, but also has the viewer's label information injected into its structure. According to the traditional concept of the discriminator in GANs, the discriminator needs to deal with both the ``fake image" case and the ``real image" case. For our network, in the case of ``real image", the image patch fed into the input layer of the discriminator is an image combining the input that we injected to the first layer of the generator and the ground truth personalized fixation heat map. In the case of a ``fake image", the ground truth personalized fixation map will be replaced by the output image generated by the generator.  In the architecture described above, the generator has the visual stimuli, non-personalized version of gaze fixations and label information mixed together as its input. Therefore, by conditioning on the label and with the help of the discriminator, the generator learns to generate more personalized saliency predictions shaped by viewers' identities.

\section{Related Work}
\label{relatedwork}

Many issues in computer vision and image generation can be represented by problems consisting in translating an input image into an output image while meeting some specified requirements. In the literature, it can be found that automatic generation of desirable output images can be accomplished with deep learning techniques given enough training data. Taking convolutional neural nets (CNNs) as an example, by telling the network to minimize a designing loss, it can generate output images that become more and more similar to the ground truth images \cite{zhang2016colorful,pathak2016context}. A generative model that is solely based on CNN often has the shortcoming of producing blurred results, caused by the  CNN minimizing a loss function by averaging all plausible outputs \cite{isola2016image}. Generative Adversarial Networks (GANs), a type of more sophisticated generative models, can satisfy the goal of producing less blurring and more realistic images \cite{goodfellow2014generative}. While training a generator along with training a discriminator that learns to distinguish a real image from a fake image at the same time, the quality of the output images can be greatly improved. Deep Convolutional GANs (DCGAN) improve upon the original GAN structure, by replacing pooling layers with convolutional layers and implementation of batch normalization. The DCGAN can generate high-resolution images \cite{radford2015unsupervised}. Since the main focus of our study is to implement a network that can model attention according to the observers' characteristics, it would be highly desirable if the network can receive as input some extra label information indicating the groups that the observers belong to and generate output images accordingly to the label information. This leads us to using Conditional GANs, which are able to use extra conditional information in the input and generate more specific samples, as our basic network \cite{mirza2014conditional}. A conditioning label is fed into the network as input. The model is then trained to generate images while having both the generator and discriminator conditioned on this label. In this way, the network learns to exploit both the input pixels and the data y when generating output images. An improved version of the Conditional GAN approach is the StackGAN \cite{zhang2016stackgan}, which is constructed by pipelining two GAN structures. The first GAN generates images with low-resolution. The second GAN acts as a refinement process. It takes the results generated from the first GAN and generates images with more details and higher resolution by making use of the input data y. It is worth noting that the second stage GAN, which uses the low-resolution images and the label information together in its generator and discriminator, is an efficient strategy to produce higher resolution images which meet the requirement dictated by the label information.

\section{Architecture and Training}

When eye tracking data are collected from observers looking at various image stimuli, their viewing behaviors are mainly affected by the different visual stimuli that are presented to them. However, the behaviors are also modulated by participants' personal characteristics. Thus, as our goal is to enable our network to predict salient regions that are personalized for different viewer categories, instead of training the network using solely the stimuli pixels, we also need to consider the complementary information, which is the label information indicating viewers' characteristics. In the case where the ground truth heat maps containing the eye fixation data of all population are available, we can benefit from this information for training, since we assume that the final personalized heat maps will be a fine-tuned version of the population fixation heat maps. The architecture of our network contains a generator and a discriminator, following the concept of GANs. More specifically, we adapted our network from the Conditional GAN and StackGAN architectures. Details of the architecture and key features will be discussed below.

\subsection{Generator}

\begin{figure}[t]
\begin{center}
\fbox{
   \includegraphics[width=0.9\linewidth,height= 16cm]{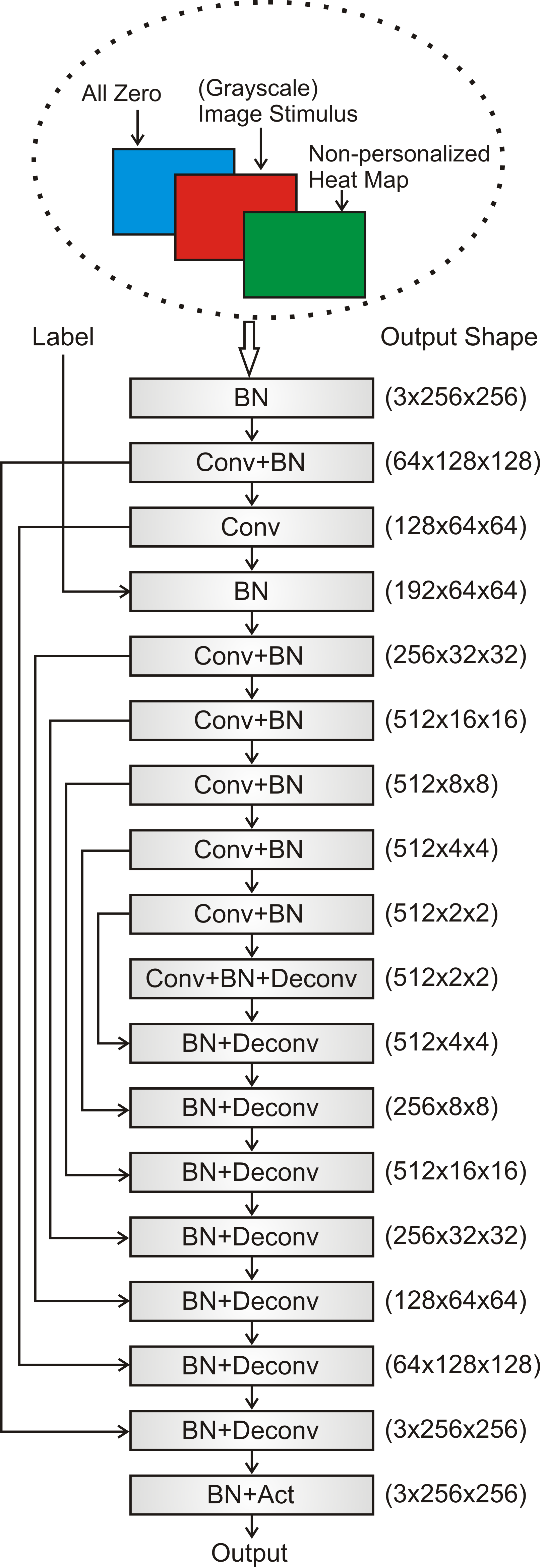}}
\end{center}
   \caption{Composition of the generator network. The generator ends with a Tanh activation (Act). In this way, we use the output of the Tanh activation layer to generate personalized saliency predictions.}
\label{fig:archi_gen}
\end{figure}

\begin{figure}[t]
\begin{center}
\fbox{
   \includegraphics[width=0.9\linewidth]{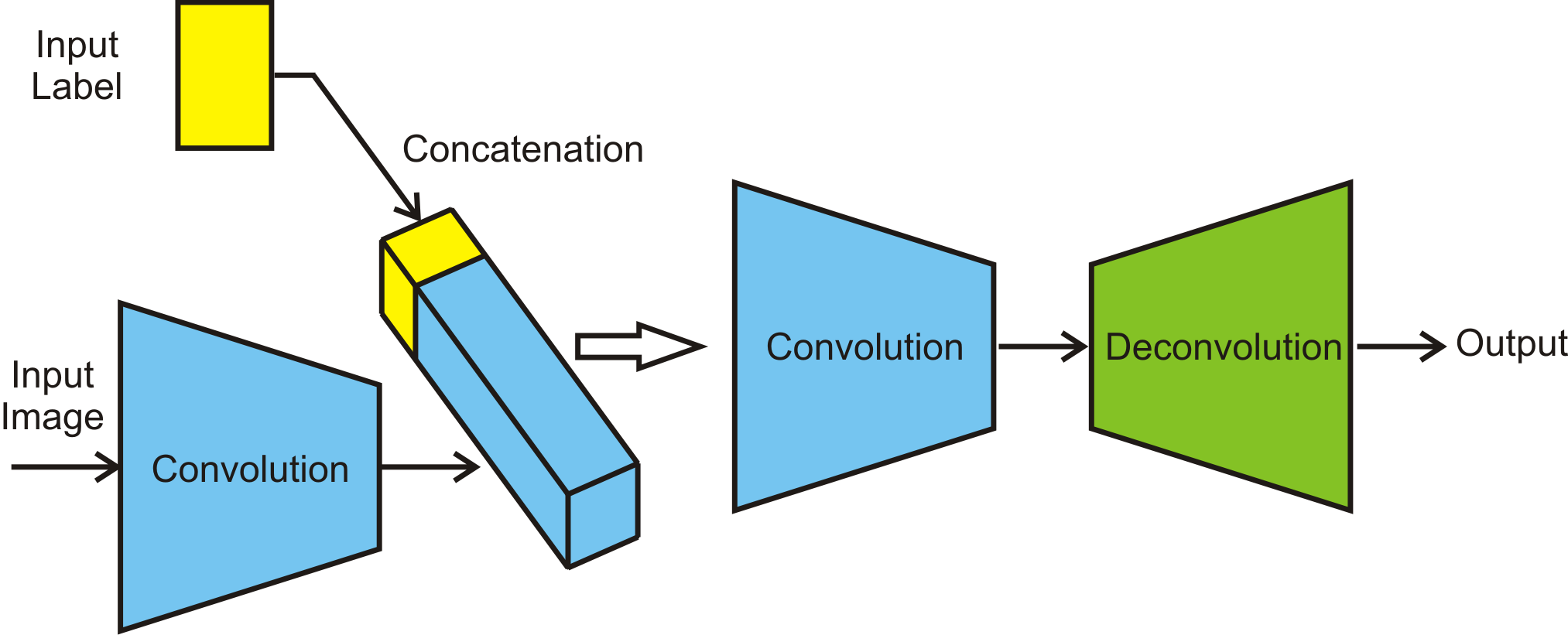}}
\end{center}
   \caption{Overall architecture of the generator.}
\label{fig:generator_draw}
\end{figure}

As shown in Figure~\ref{fig:archi_gen}, our generator has an encoder-decoder architecture. The encoder part consists of modules of the form Convolution-Relu-BatchNormalization. All convolutions use ReLU activations to provide non-linearity for our network. Batch normalization is applied after every convolutional layer \cite{ioffe2015batch} in order to accelerate training. The generator takes two inputs.  One input is an RGB image of size (3 x 256 x 256), with the R channel encoding the (grayscale) image stimulus, the G channel encoding the population fixation heat map and the B channel set to all zero. We found that combining the input elements in this way improved the training over just simple concatenation. This RGB image is fed into the input convolutional layer of the encoder. The other input is the label information representing the viewer groups (e.g. young vs. old or native language vs. second language depending on the specific application). The (binary) label is extended by simple copying to all elements in a (64 x 64 x 64) tensor, which is injected into the interior of the network by concatenation with the 64x64 level of the encoder. This spatial dimension is determined based on empirical observation. If we use a smaller spatial dimension and inject the label information too early into the generator, the network will have difficulty in training the first few layers of the network, since these layers are supposed to extract some low-level features from the image stimuli independently of the label information. However, it is also inadequate to inject the label information too late. It has been observed that injecting the label information into the last convolutional layer of the generator will inhibit the generator from including the label information during its encoding process. Since we focus here only on the Age Study and GECO datasets, we chose to solely consider two possibilities when encoding the label information. A (64 x 64 x 64) dimensional tensor with all zeros represents the younger viewer group for the Age Study and a tensor with all ones represents the elder group. As for the GECO dataset, a (64 x 64 x 64) dimensional tensor with all zeros represents the native language reading case, and a tensor with all ones represents the non-mother tongue reading. Our input RGB image is fed into the encoder through a series of convolutions until it has a size suitable for being concatenated with the (64 x 64 x 64) label information tensor. After the last layer of the encoder, the feature maps are of the size (512 x 1 x 1). The decoder part of the generator consists of modules of the form Deconvolution-ReLU-BatchNormalization. Five dropout layers, which randomly drop 20 percent of the input units, are inserted after the batch normalization layers to improve learning. It should be noted that we also implemented the U-Net architecture mentioned in \cite{isola2016image}. Mirrored layers in the encoder and decoder are merged via our decoder implementation, and the activations from those layers are connected during training. The generator ends with a Tanh activation. In this way, we use the output of the Tanh activation layer to generate the personalized saliency prediction. Figure~\ref{fig:generator_draw} provides an illustration of the generator architecture.

\subsection{Discriminator}



\begin{figure}[t]
\begin{center}
\fbox{
   \includegraphics[width=0.9\linewidth,height= 9cm]{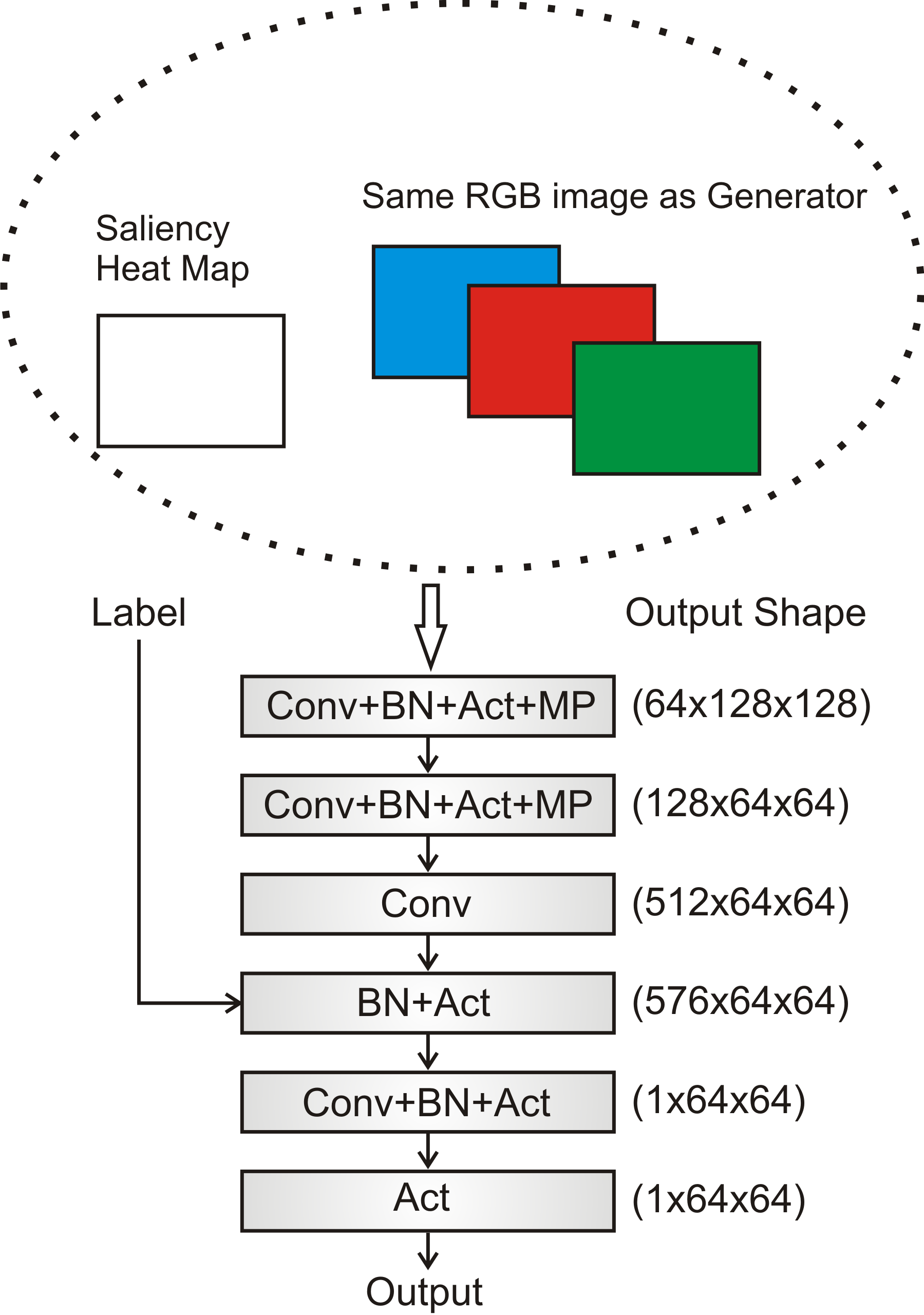}}
\end{center}
   \caption{Composition of the discriminator network. Two MaxPooling (MP) layers are used in the first two convolution modules to reduce the size of the output. All the activation (Act) functions, except the last one, consist of Tanh activations. The output of the last convolutional layer is followed by a sigmoid activation (Act) to produce the final output. }
\label{fig:archi_dis}
\end{figure}

\begin{figure}[t]
\begin{center}
\fbox{
   \includegraphics[width=0.9\linewidth]{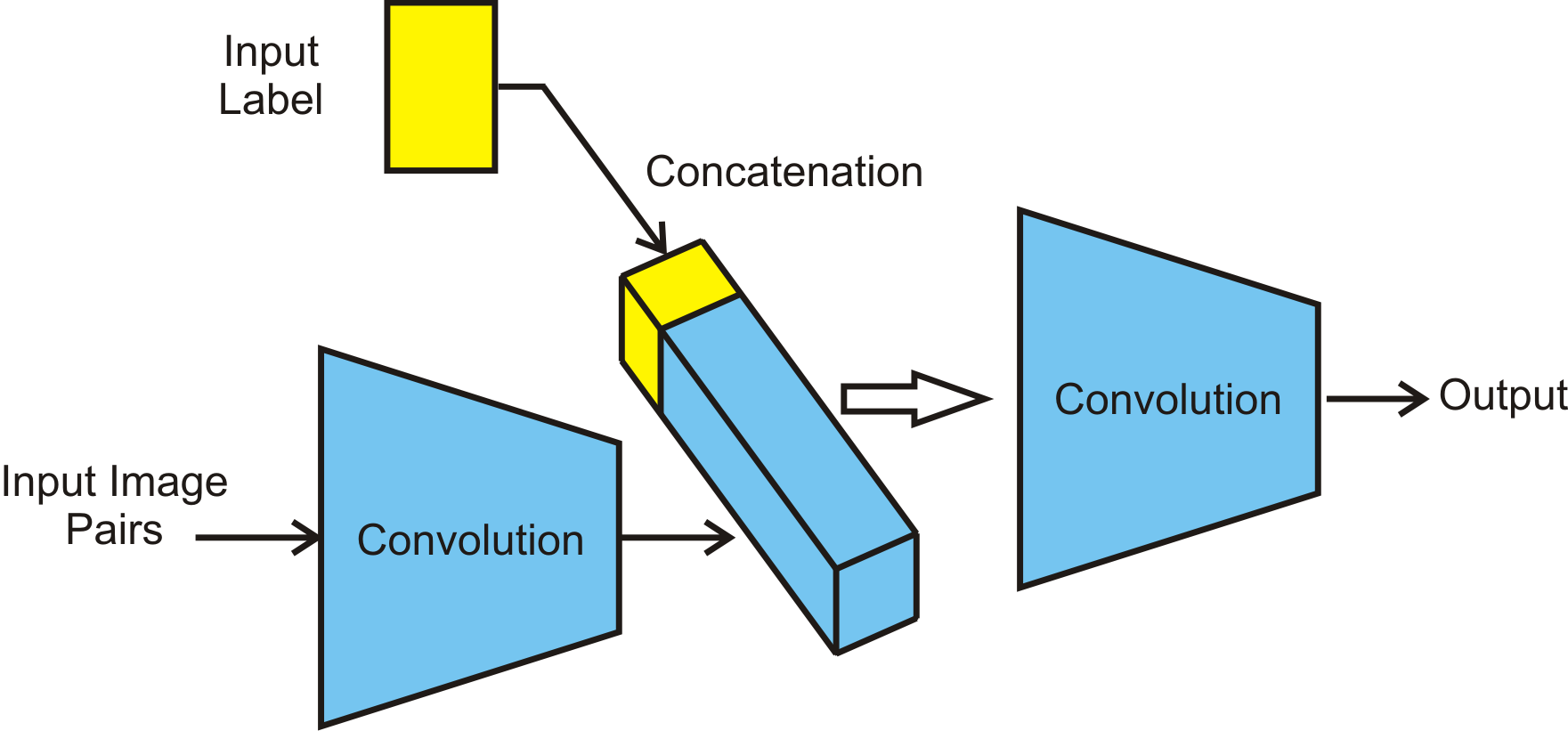}}
\end{center}
   \caption{Overall architecture of the discriminator.}
\label{fig:discriminator_draw}
\end{figure}

As shown in Figure~\ref{fig:archi_dis}, our discriminator consists of several layers that mainly implement four convolution modules in sequence. The first two modules are of the form Convolution-BatchNormalization-Activation (Tanh)-MaxPooling. Two pooling layers of pooling size (2 x 2) are used to reduce the size of the output. Then the last two modules are of the form Convolution-BatchNormalization-Activation (Tanh). Finally, the output of the last convolutional module is followed by a sigmoid activation, which allows the discriminator to produce its output result. The discriminator is also fed with two inputs. At its input layer, it receives as input an RGB image. It consists of the RGB pixels fed to the generator concatenated with a saliency heat map. According to the concept of ``real" and ``fake" saliency map cases for the discriminator, the ``fake" case corresponds to the situation when the prediction generated by the generator is chosen as the saliency heat map. As for the ``real" case, we chose the ground truth personalized fixation heat map as the concatenated saliency heat map. The other input is the (64 x 64 x 64) label information tensor. It is merged with the third module of the discriminator, which has an output shape that is suitable for being concatenated with the (64 x 64 x 64) label tensor.
Figure~\ref{fig:discriminator_draw} provides an illustration of the discriminator architecture.

\subsection{Training}

The proposed network is implemented using Keras. It is well known that Conditional GANs, different than GANs  which learn to generate the desirable output image from a random noise vector, are trained to learn a mapping from the random noise vector combined with a complementary input to the desired output image. Furthermore, as stated in \cite{isola2016image}, when doing image translation using Conditional GAN, the random noise vector can be removed and replaced simply by dropout layers in the architecture. Thus we chose to implement our network and the loss functions without considering the random noise vector. In this way, the discriminator, whose task is to classify between real and fake pairs, uses the following binary cross entropy loss as its loss function: 

\begin{multline}
L_D = E_{x,v,y}[log D(x,v,y) ] +\\
E_{x,v}([1-log D(x,v,G(x,v)) ]
\label{equ:disc}
\end{multline}

In Equation~\ref{equ:disc}, $v$ is the user label, $x$ in the input RGB, $y$ is the real personalized ground truth heat map. As for the generator, since it is stated in \cite{pathak2016context} that mixing the GAN adversarial loss with another standard content loss such as Euclidean loss can improve training of deep neural networks, we choose to use the $L_1$ distance as the additional loss and combine it with the adversarial loss described by Equation~\ref{equ:disc} to construct the overall loss function for our generator. The $L_1$ distance is designed to measure the difference between the generator's output and the ground truth personalized fixations. Therefore, the overall loss function for our generator is defined as:

\begin{equation}
L_G = L_D + \lambda L_1(G)
\label{equ:gen}
\end{equation}

We set the value of $\lambda$  to 0.01, based on our observation from experiments and on the analysis made in \cite{isola2016image}, which indicates that, when the $L_1$ loss is weighted 100 times larger than the GAN loss, there are fewer artifacts produced as the output of the generator. All layers of the network need to be trained from scratch. Weights are randomly initialized using a uniform distribution between $-0.05$ to $0.05$. We always reserve 20 percent of the total images for testing. The network is trained by updating the weights of the generator and the discriminator in alternation. The GAN cross-entropy loss is backpropagated to the discriminator to update its weights. Then, by keeping the discriminator weights constant, we combine the cross-entropy loss with the $L_1$ loss and backpropagate this error to update the generator weights. Minibatches of two samples per batch is used for training. The RMSProp optimizer is used to optimize both the generator and the discriminator, with a learning rate of 0.001, a decay rate of 0.9, a momentum of 0 and an $\epsilon$ of $1\times10^{-6}$. Dropout layers and batch normalization are used in our network to accelerate convergence as mentioned above. 

\section{Evaluation and Discussion }

\begin{table}
\begin{center}
\begin{tabular}{|l|c|c|}
\hline
 & KL& SSIM \\
\hline\hline
\shortstack[l]{All population eye fixations  \\against ground truth for seniors } & 3.84& 0.52 \\
\hline
Saliency personalized for seniors & 2.49& 0.62 \\
\hline
\shortstack[l]{All population eye fixations\\ against ground truth for juniors} & 6.00& 0.47\\
\hline
Saliency personalized for juniors&2.27&0.66\\
\hline
\end{tabular}
\end{center}
\caption{KL-divergence and SSIM scores for the prediction performance of eye fixations including all population vs. personalized saliency predictions.}
\label{table:age_popvspersonalized}
\end{table}

\begin{table}
\begin{center}
\begin{tabular}{|l|c|c|c|c|}
\hline
 & AUC& NSS &KL&SSIM \\
\hline\hline
MLNET (S) & 0.73& 0.90 &8.75&0.25\\
Personalized saliency (S)& 0.74& 1.08&6.73&0.41 \\
MLNET (J) & 0.75& 1.08&10.69&0.24\\
Personalized saliency (J)&0.76&1.15&8.13&0.47\\
\hline
\end{tabular}
\end{center}
\caption{Evaluation scores for MLNET saliency predictions vs. personalized predictions generated using our model on test sets. The scores for the saliency predictions applied on the senior group (S) are presented in the first two rows. The scores for the saliency predictions applied on the junior group (J) are presented in the last two rows. }
\label{table:AUCNSSKLSIM}
\end{table}

\begin{table}
\begin{center}
\begin{tabular}{|l|c|c|}
\hline
 & MSE \\
\hline\hline
Non-Personalized Saliency (L1) & 2.55\\
Non-Personalized Saliency (L2) & 2.53\\
Personalized Saliency (L1) & 1.70 \\
Personalized Saliency (L2) & 1.71 \\

\hline
\end{tabular}
\end{center}
\caption{MSE evaluation scores obtained on test sets. The first two rows show the results obtained from the comparison between non-personalized eye fixations and the ground truth for both the native reader (L1) and second language reader (L2) cases. The last two rows show the results obtained from the comparison between the generated saliency heat maps and the ground truth for both L1 and L2 cases.}
\label{table:msefortext}
\end{table}

When our personalized saliency model is applied to the Age Study dataset, since our generated output images need to be compared against our ground truth saliency heat maps, we considered several evaluation metrics that are the methods of choice in the saliency modeling literature. Among them, we selected four popular evaluation metrics: Area Under the ROC Curve (AUC), Normalized Scan-path Saliency (NSS), KL-divergence (KL) and Structure Similarity (SSIM) \cite{bylinskii2016different,wang2004image}.

When the Age Study dataset is used to train our saliency personalization model, general eye fixation heat maps from the overall population are fed into the network as input, and the network is expected to produce age-related grayscale saliency predictions. Therefore, we used KL and SSIM, which are reported in Table~\ref{table:age_popvspersonalized}, to demonstrate the ability of our model to generate personalized saliency maps accordingly to the observers' age information. The results show that the personalized saliency maps provide an improved prediction for both age groups. Evaluated against the ground truth age-specified eye fixations, the prediction performance of our personalized saliency maps consistently outperforms the performance of the eye fixation heat maps involving the overall population.

Although the ground truth eye fixations involving the overall population are fed into our network as input during training, for the post-training application of the network typically we will not have the actual (population) heat map. Therefore, in attempt to test the usefulness of our design and verify whether the training result is applicable for solving practical saliency prediction issues, during the testing phase, we feed our network with the saliency heat maps produced by MLNET as the input population eye fixation ground truth. We  chose the MLNET saliency model as it was (at time of writing) one of the state-of-the-art deep learning based salience methods. Other methods could also be used, as long as they are trained on databases representative of the population. Table~\ref{table:AUCNSSKLSIM} compares the prediction performance of our personalized saliency model with the MLNET saliency map as input using AUC, NSS, KL-divergence and Similarity metrics respectively. The results show that our network successfully produces saliency predictions that are personalized with observers' characteristics, and the improvements of saliency prediction accuracy are observed in both the senior and junior age groups.

\begin{figure}[t]
\begin{center}
\fbox{
   \includegraphics[width=0.9\linewidth]{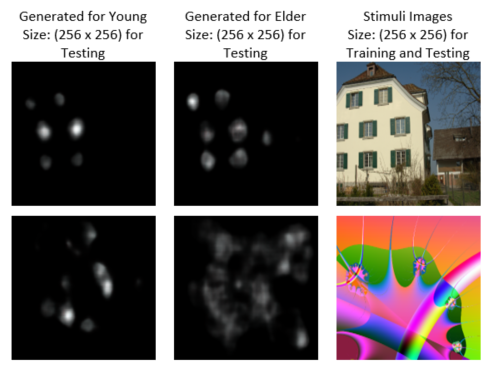}}
\end{center}
   \caption{Saliency maps generated from our model. From the left, the first column is the output of our network when taking image stimuli along with the label indicating ``younger viewer group" as inputs. The second column is the output generated when the injected label indicates ``elder viewer group". The last column presents the corresponding image stimuli. We set all the heat maps and stimuli images to the size of (256 x 256) for the training and testing of our network.  }
\label{fig:person_output_YandO}
\end{figure}

\begin{figure}[t]
\begin{center}
\fbox{
   \includegraphics[width=0.9\linewidth]{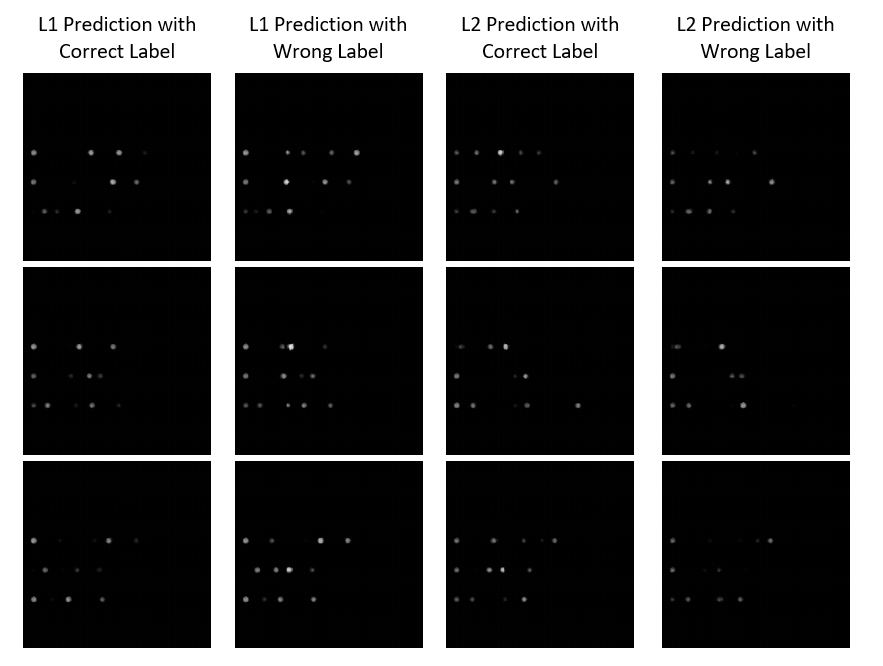}}
\end{center}
   \caption{ Saliency heat maps generated by using four different types of inputs. From the left, the first column is the results generated when images containing Dutch words along with label ``L1"  are fed into network. The second column is produced when images containing Dutch words are fed into network along with a wrong label indicating ``L2". The third column is produced when using images containing English material along with a correct label indicating ``L2". The fourth column is produced when using images containing English material along with a wrong label indicating ``L1". }
\label{fig:l1l2outputdifflabels}
\end{figure}

To investigate the significance of our contribution to personalizing saliency predictions, we examined the generated outputs belonging to different age groups in a more visual way. By human inspection, it can be seen clearly in Figure~\ref{fig:person_output_YandO} that the saliency predictions generated for the younger observer group are fine-tuned to be more condensed and centralized. However, the saliency predictions generated for the elder observer group are more spread out and smooth. Those observations are consistent with the findings reported in the Age Study paper \cite{accik2010developmental}. 

To evaluate the performance of our network to predict eye fixations during text reading for the GECO dataset, we used the standard metric, MSE, to evaluate the prediction performance of our generated eye fixation heat maps. In Table~\ref{table:msefortext}, the first two rows show the results obtained by comparing the non-personalized eye fixation heat maps against the ground truth for both the native speaker (L1) and second language (L2) reading contexts. These scores are obtained by assuming that the readers land their eyes uniformly at the center location of each word presented in an image. By comparing these scores with the scores presented in the third and fourth rows, we see that our model is able to produce personalized saliency heat maps with higher prediction accuracy. Also, it can be observed in Figure~\ref{fig:l1l2outputdifflabels} that when our network is fed with label L1, the predicted saliency heat maps show a viewing pattern with higher skipping rate; however, when fed with label L2, there are more fixation points and they are more spread out over the images. This observation is in agreement with the conclusions stated in \cite{cop2017presenting}, thus this shows that the saliency predictions produced by our network are well adjusted to the input label. 

\section{Conclusion}

We developed a Deep Convolutional Generative Adversarial Network (DCGAN) for generating personalized saliency predictions without any extra information collection or data construction operations. Our model combines user-specific information with image stimuli to predict gaze patterns that can be adjusted to various observers' backgrounds and viewing contexts. Both quantitative and qualitative results demonstrated the effectiveness of our proposed approach for producing highly accurate saliency predictions in the case of personalization based on age or native language. The personalized training result of the network could be useful for other related tasks such as personality classification and language aptitude evaluation. Since the GAN approach and its derivatives have been extensively explored in the past few years in the field of computer vision, we believe that the introduction of personalized features into the GAN-based model, along with the adaptability of our model to a variety of viewing contexts, will bring more research possibilities in the attention field and lead to a greater range of application areas.

{\small
\bibliographystyle{ieee}
\bibliography{egbib}
}

\end{document}